\documentclass[10pt,twocolumn,letterpaper]{article}

\usepackage{cvpr}
\usepackage{times}
\usepackage{epsfig}
\usepackage{graphicx}
\usepackage{amsmath}
\usepackage{amssymb}
\usepackage{multirow}

\usepackage[breaklinks=true,bookmarks=false]{hyperref}

\cvprfinalcopy 

\def\cvprPaperID{2289} 

\begin{document}

\title{Scene Graph Generation by Iterative Message Passing}

\author{Danfei Xu\textsuperscript{1}\quad Yuke Zhu\textsuperscript{1}\quad Christopher B. Choy\textsuperscript{2}\quad Li Fei-Fei\textsuperscript{1}
	\\ \textsuperscript{1}Department of Computer Science, Stanford University
	\\ \textsuperscript{2}Department of Electrical Engineering, Stanford University
	\\ {\tt\small \{danfei, yukez, chrischoy, feifeili\}@cs.stanford.edu }
}

\maketitle
\begin{abstract}
Understanding a visual scene goes beyond recognizing individual objects in isolation. Relationships between objects also constitute rich semantic information about the scene. In this work, we explicitly model the objects and their relationships using scene graphs, a visually-grounded graphical structure of an image. We propose a novel end-to-end model that generates such structured scene representation from an input image. The model solves the scene graph inference problem using standard RNNs and learns to iteratively improves its predictions via message passing. Our joint inference model can take advantage of contextual cues to make better predictions on objects and their relationships. The experiments show that our model significantly outperforms previous methods for generating scene graphs using Visual Genome dataset and inferring support relations with NYU Depth v2 dataset.
\end{abstract}


\section{Introduction}
\label{sec:intro}

\begin{figure}[t!]
\begin{center}
\includegraphics[width=1.0\linewidth]{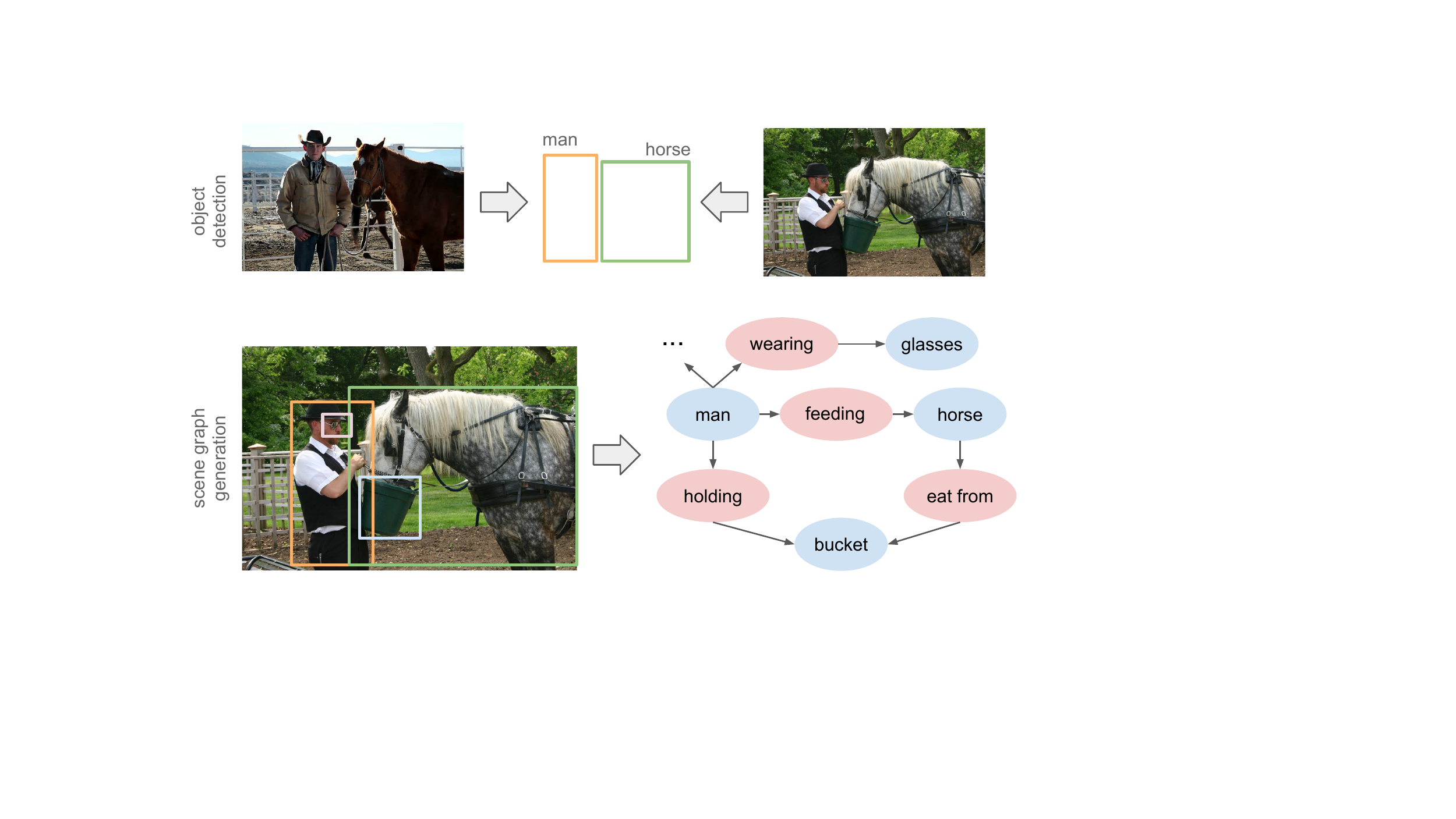}
\caption{Object detectors perceive a scene by attending to individual objects. As a result, even a perfect detector would produce similar outputs on two semantically distinct images (first row). We propose a scene graph generation model that takes an image as input, and generates a visually-grounded scene graph (second row, right) that captures the objects in the image (blue nodes) and their pairwise relationships (red nodes).}
\vspace{-20pt}
\end{center}
\label{fig:intro_figure}
\end{figure}

Today's state-of-the-art perceptual models~\cite{he2015deep,renNIPS15fasterrcnn} have mostly tackled detecting and recognizing individual objects \emph{in isolation}. However, understanding a visual scene often goes beyond recognizing individual objects. Take a look at the two images in Fig.~\ref{fig:intro_figure}. Even a perfect object detector would struggle to perceive the subtle difference between a man feeding a horse and a man standing by a horse. 
The rich semantic relationships between these objects have been largely untapped by these models. 
As indicated by a series of previous works~\cite{lu2016visual,sadeghi2011recognition,zhao2013scene}, one crucial step towards a deeper understanding of visual scenes is building a structured representation that captures objects and their semantic relationships.
Such representation not only offers contextual cues for fundamental recognition tasks~\cite{mottaghi2014role,oliva2007role,torralba2003contextual,yao2010modeling} but also provide values in a larger variety of high-level visual tasks~\cite{Johnson2015-lm,zitnick2013learning,yatskarsituation}.

The recent success of deep learning-based recognition models~\cite{he2015deep,krizhevsky2012imagenet,simonyan2014very} has surged interest in
examining the detailed structures of a visual scene, especially in the form of
object relationships~\cite{chao2015hico,VG,lu2016visual,BMVC2015_52}. Scene graph, proposed by Johnson \etal~\cite{Johnson2015-lm}, offers a platform to explicitly model objects and their relationships. In short, a \emph{scene graph} is a visually-grounded graph over the object instances in an image, where the edges
depict their pairwise relationships (see example in Fig.~\ref{fig:intro_figure}). 
The value of scene graph representation has been proven in a wide range of visual tasks, such as semantic image retrieval~\cite{Johnson2015-lm}, 3D scene synthesis~\cite{chang2014learning}, and visual question answering~\cite{teney2016graph}. Anderson \etal recently proposed SPICE~\cite{spice2016} as an enhanced automated caption evaluation metric defined over scene graphs.
However, these models that use scene graphs either rely on
ground-truth annotations~\cite{Johnson2015-lm}, synthetic images~\cite{teney2016graph}, or extract a scene graph from text domain~\cite{spice2016,chang2014learning}. 
To truly take advantage of such rich structure, it is crucial to devise a model that automatically generates scene graphs from images. 

In this work, we address the problem of scene graph generation, where the goal is to generate a visually-grounded scene graph from an image. In a generated scene graph, an object instance is characterized by a bounding box with an object category label, and a relationship is characterized by a directed edge between two bounding boxes (i.e., object and subject) with a relationship predicate (red nodes in Fig.~\ref{fig:intro_figure}).
The major challenge of generating scene graphs is reasoning about relationships. Much effort has been expended on localizing and recognizing semantic relationships in images~\cite{HICO,desai2010discriminative,lu2016visual,sadeghi2011recognition,yao2010modeling}. Most methods have focused on making \emph{local} predictions of object relationships~\cite{lu2016visual,sadeghi2011recognition}, which essentially simplify the scene graph generation problem into independently predicting relationships between pairs of objects. However, by doing \emph{local} predictions these models ignore surrounding context, whereas joint reasoning with contextual information can often resolve ambiguity due to local predictions in isolation.

\begin{figure}[htp!]
    \begin{center}
        \includegraphics[width=1.0\linewidth]{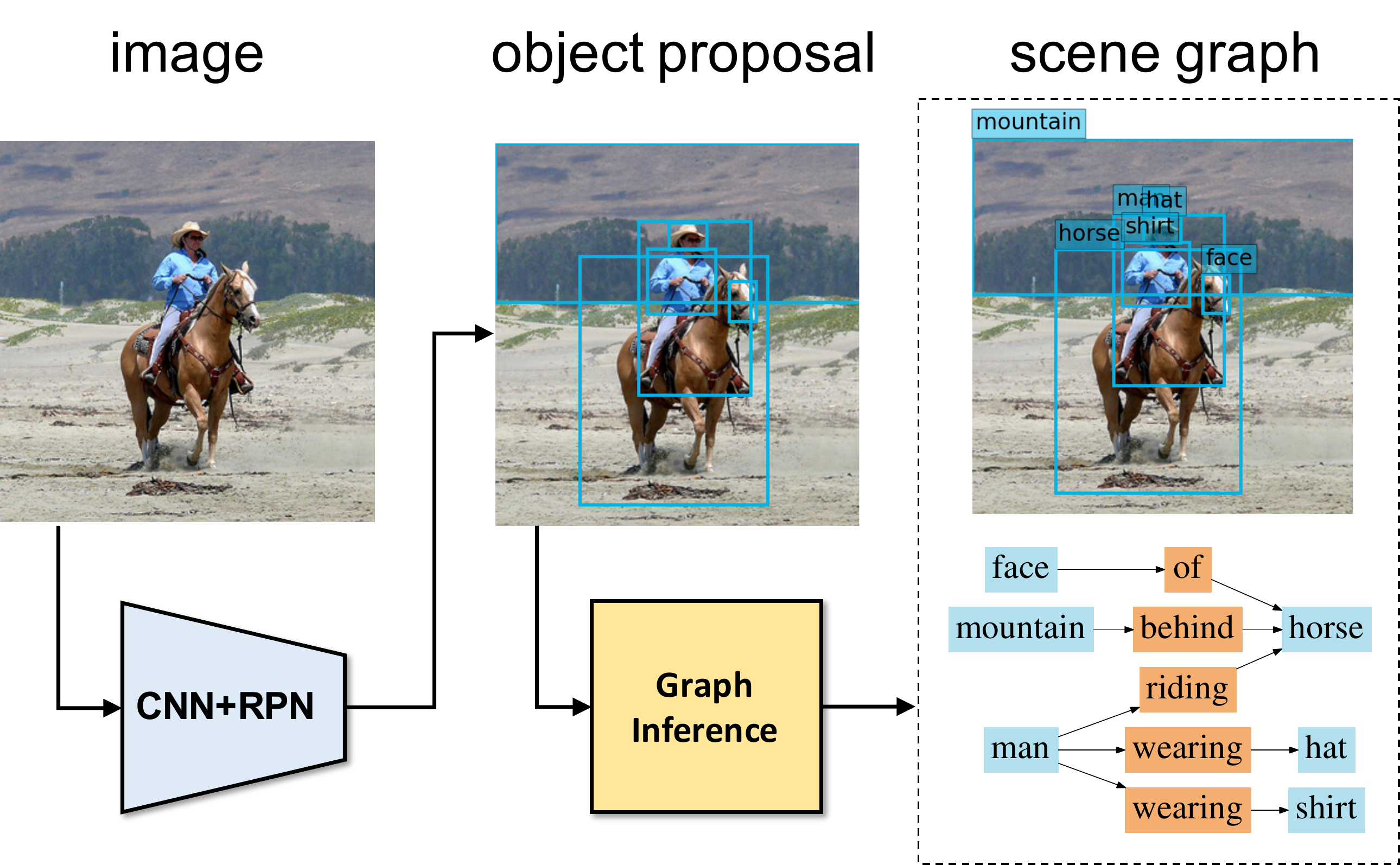}
\vspace{3pt}
        \caption{An overview of our model architecture. Given an image as input, the model first produces a set of object proposals using a Region Proposal Network (RPN)~\cite{renNIPS15fasterrcnn}, and then passes the extracted features of the object regions to our novel graph inference module. The output of the model is a \emph{scene graph}~\cite{Johnson2015-lm}, which contains a set of localized objects, categories of each object, and relationship types between each pair of objects.}
    \label{fig:pipeline}

    \end{center}
\vspace{-10pt}

\end{figure}

To capture this intuition, we propose a novel end-to-end model that learns to generate image-grounded scene graphs (Fig.~\ref{fig:pipeline}). The model takes an image as input and outputs a scene graph that consists of object categories, their bounding boxes, and semantic relationships between pairs of objects. Our major contribution is that instead of inferring each component of a scene graph in isolation, the model passes messages containing contextual information between a pair of bipartite sub-graphs of the scene graph, and iteratively refines its predictions using RNNs.
We evaluate our model on a new scene graph dataset based on Visual
Genome~\cite{VG}, which contains human-annotated scene graphs on 108,077 images. On average, each image is annotated with 25 objects and 22 pairwise object relationships. We show that relationship prediction in scene graphs can be significantly improved by our model. Furthermore, we also apply our model to the NYU Depth v2 dataset~\cite{nyudepth}, establishing new state-of-the-art results in reasoning about spatial relations, such as horizontal and vertical supports.

In summary, we propose an end-to-end model that generates visually-grounded scene graphs from images. The model uses a novel inference formulation that iteratively refines its prediction by passing contextual messages along the topological structure of a scene graph. We demonstrate its use for generating semantic scene graphs from a new scene graph dataset as well as predicting support relations using the NYU Depth v2 dataset~\cite{nyudepth}.

\section{Related Work}

\noindent \textbf{Scene understanding and relationship prediction.}
Visual scene understanding often harnesses the statistical patterns of object co-occurrence~\cite{galleguillos2008object,
ladicky2010graph, rabinovich2007objects, salakhutdinov2011learning} as well as spatial
layout~\cite{baur2008statistics, desai2011discriminative}. 
A series of contextual models based on surrounding pixels and regions have also been developed for perceptual tasks~\cite{bell2015inside,RCNN,lin2013holistic,mottaghi2014role}.
Recent works~\cite{HICO,ramanathan2015learning} exploits more complex structures for relationship prediction. However, these works focus on image-level predictions without detailed visual grounding.
Physical relationships, such as support and stability, have been studied in~\cite{Jia2013-la, nyudepth, Zheng2015-tm}. 
Lu \etal \cite{lu2016visual} directly tackled the semantic relationship detection by combining visual inputs with language priors to cope with the long-tail distribution of real-world relationships. However, their method predicts each relationship independently. We show that our model outperforms theirs with joint inference.
\vspace{5pt}

\noindent \textbf{Visual scene representation.}
One of the most popular ways of representing a visual scene is through text descriptions~\cite{gupta2008beyond,sadeghi2011recognition,zitnick2013learning}. Although text-based representation has been shown to be helpful for scene classification and retrieval, its power is often limited by ambiguity and lack of expressiveness.
In comparison, scene graphs~\cite{Johnson2015-lm} offer explicit grounding of visual concepts, avoiding referential uncertainty in text-based representation. 
Scene graphs have been used in many downstream tasks such as image retrieval~\cite{Johnson2015-lm}, 3D scene synthesis~\cite{chang2014learning} and understanding~\cite{Fisher2011-ci}, visual question answering~\cite{teney2016graph}, and automatic caption evaluation~\cite{spice2016}. 
However, previous work on scene graphs shied away from the graph generation problem by either using ground-truth annotations~\cite{Johnson2015-lm,teney2016graph}, or extracting the graphs from other modalities~\cite{spice2016,chang2014learning,Fisher2011-ci}. Our work addresses the problem of generating scene graphs directly from images.

\vspace{5pt}

\noindent \textbf{Graph inference.} 
Conditional Random Fields (CRF) have been used extensively in graph inference. Johnson \etal used CRF to infer scene graph grounding distributions for image retrieval~\cite{Johnson2015-lm}.  Yatskar \etal~\cite{yatskarsituation} proposed situation-driven object and action prediction using a deep CRF model. Our work is closely related to CRFasRNN~\cite{crfasrnn} and Graph-LSTM~\cite{glstm} in that we also formulate the graph inference problem using an RNN-based model. A key difference is that they focus on node inference while treating edges as pairwise constraints, whereas we enable  edge predictions using a novel primal-dual graph inference scheme.
We also share the same spirit as Structural RNN~\cite{jain2015structural}. A crucial distinction is that our model iteratively refines its predictions through message passing, whereas the Structural RNN model only makes one-time predictions along the temporal dimension, and thus cannot refine its past predictions.  

\section{Scene Graph Generation}
\label{sec:methods}

\begin{figure*}[!htp]
  \begin{center}
    \includegraphics[width=\linewidth]{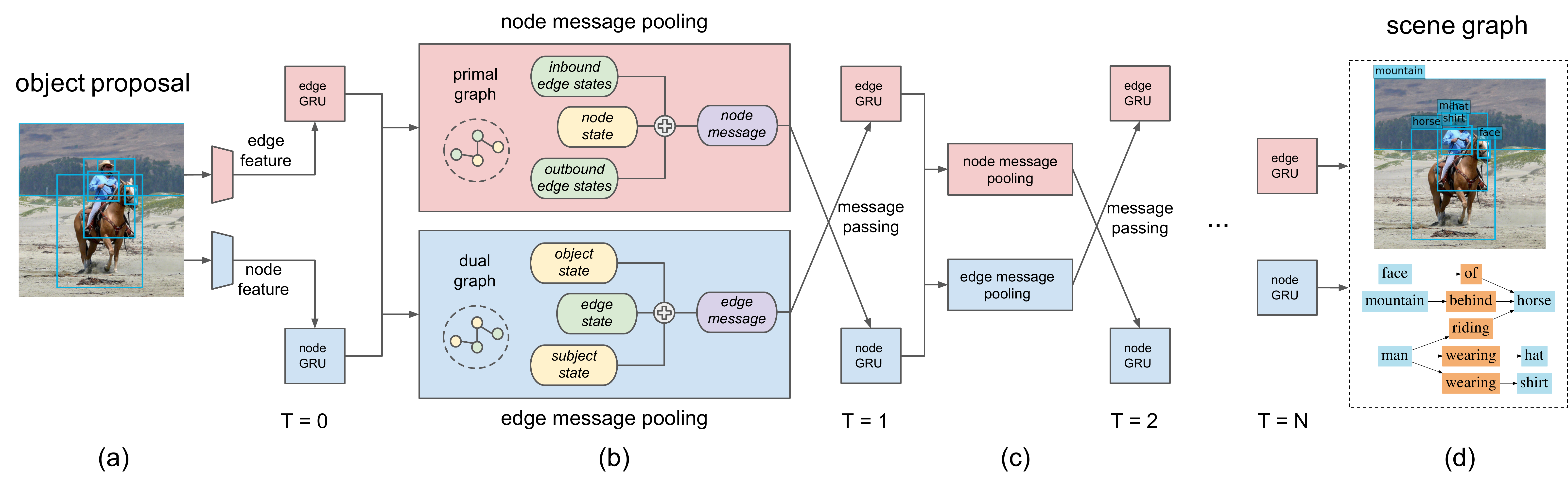}
 \vspace*{-15pt}

  \end{center}
   \caption{An illustration of our model architecture (Sec.~\ref{sec:methods}). The model first extracts visual features of nodes and edges from a set of object proposals, and edge GRUs and node GRUs then take the visual features as initial input and produce a set of hidden states (a). Then a node message pooling function computes messages that are passed to the node GRU in the next iteration from the hidden states.  Similarly, an edge message pooling function computes messages and feed to the edge GRU (b). The $\oplus$ symbol denotes a learnt weighted sum. The model iteratively updates the hidden states of the GRUs (c). At the last iteration step, the hidden states of the GRUs are used to predict object categories, bounding box offsets, and relationship types (d).}
  \label{fig:architecture}
\end{figure*}

A \emph{scene graph}, as defined by Johnson \etal~\cite{Johnson2015-lm}, is a structured representation of an image, where nodes in a scene graph correspond to object bounding boxes with their object categories, and edges correspond to their pairwise relationships between objects. 
The task of \emph{scene graph generation} is to generate a visually-grounded scene 
graph that most accurately correlates with an image.
Intuitively, individual predictions of objects and relationships can benefit from their surrounding context.
For instance, knowing ``a horse is on grass field'' is likely to increase the chance of detecting a person and predicting the relationship of ``man riding horse''.
To capture this intuition, we propose a joint inference framework to enable contextual information to propagate through the scene graph topology via a message passing scheme.

Inference on a densely connected graph can be very expensive. As shown in previous work~\cite{dense_crf} and ~\cite{crfasrnn}, dense graph inference can be approximated by mean field in Conditional Random Fields (CRF). Our approach is inspired by Zheng~\etal~\cite{crfasrnn}, which designs fully differentiable layers to enable end-to-end learning with recurrent neural networks (RNN). Yet their model relies on purpose-built RNN layers. To achieve greater flexibility in a more principled training framework, we use a generic RNN unit instead, in particular a Gated Recurrent Unit (GRU)~\cite{gru}. At each iteration, each GRU takes its previous hidden state and an incoming message as input, and produces a new hidden state as output. Each node and edge in the scene graph maintains its internal state in its corresponding GRU unit, where all nodes share the same GRU weights (node GRUs), and all edges share the other set of GRU weights (edge GRUs).  This setup allows the model to pass messages (i.e., aggregation of GRU hidden states) among the GRU units along the scene graph topology. We also propose a message pooling function that learns to dynamically aggregate the hidden states of the GRUs into messages.

We further observe that the unique structure of scene graphs forms a bipartite structure of message passing channels. Since messages only pass along the topological structure of a scene graph, the set of edge GRUs and the set of node GRUs form a bipartite graph, where no message is passed inside each set.
Inspired by this observation, we formulate two disjoint sub-graphs that are essentially the dual graph to each other. 
The primal graph defines channels for messages to pass from edge GRUs to node GRUs.
The dual graph defines channels for messages to pass from node GRUs to edge GRUs.
With such primal-dual formulation, we can therefore improve inference efficiency by iteratively passing messages between these sub-graphs instead of through a densely connected graph. Fig.~\ref{fig:architecture} gives an overview of our model.

\subsection{Problem Formulation}
We first lay out the mathematical formulation of our scene graph generation problem. To generate a visually grounded scene graph, we need to obtain an initial set of object bounding boxes. These bounding boxes can be either from ground-truth human annotation or algorithmically generated. In practice, we use the Region Proposal Network (RPN)~\cite{renNIPS15fasterrcnn} to automatically generate a set of object bounding
box proposals $B_I$ from an image $I$ as the base input to the inference procedure (Fig.~\ref{fig:architecture}(a)). 

For each object box proposal, we need to infer two types of object-centric variables: 1) an object class label, and 2) four bounding box offsets relative to the proposal box coordinates, which are used for refining the proposal boxes. In addition, we need to infer a relationship-centric variable between every pair of proposal boxes, which denotes the predicate type of the relationship between the corresponding object pair. Given a set of object classes $\mathcal{C}$ (including background) and
a set of relationship types $\mathcal{R}$ (including none relationship), 
we denote the set of all variables to be $\mathbf{x} = \{x^{cls}_i, x^{bbox}_i,  x_{i\rightarrow j} | i = 1\dots n, j=1\dots n, i\neq j\}$, where $n$ is the number of proposal boxes, $x^{cls}_i \in \mathcal{C}$ is the class label of the $i$-th proposal box, $x^{bbox}_i \in \mathbb{R}^4$ is the bounding box offsets relative to the $i$-th proposal box coordinates, and 
 $x_{i\rightarrow j} \in \mathcal{R}$ is the relationship predicate between the $i$-th and the $j$-th proposal boxes.

At the high level, the inference task is to classify objects, predict their bounding box offsets, and classify relationship predicates between each pair of objects.  Formally, we formulate the scene graph generation 
problem as finding the optimal $\mathbf{x}^{*}=\arg\max_\mathbf{x}\Pr(\mathbf{x} | I,B_I)$ that maximizes the following probability function given the image $I$ and box proposals $B_I$:

\begin{equation}
    \Pr(\mathbf{x} | I,B_I) = \prod_{i \in V} \prod_{j\neq i} \Pr(x^{cls}_i, x^{bbox}_i, x_{i\rightarrow j} | I,B_I).
\end{equation}

In the next subsection, we introduce a way to approximate the inference procedure using an iterative message passing scheme modeled with Gated Recurrent Units~\cite{gru}.

\subsection{Inference using Recurrent Neural Network}
\label{sec:rnn}

We use mean field to perform approximate inference. We denote the probability of each variable $x$ as $Q(x|\cdot)$, and assume
that the probability only depends on the current state of each node and edge at each iteration.
In contrast to Zheng~\etal~\cite{crfasrnn}, we use a generic RNN module to compute the hidden states. In particular, we choose Gated Recurrent Units~\cite{gru} due to its simplicity and effectiveness. We use the hidden state of the corresponding GRU, a high-dimensional vector, to represent the current state of each node and each edge. 
As all the nodes (edges) share the same update rule, we share the same set of parameters among all the node GRUs, and the other set of parameters among all the edge GRUs (Fig.~\ref{fig:architecture}). 
 We denote the current hidden state of node $i$ as $h_i$ and the current hidden state of edge $i\rightarrow j$ as $h_{i \rightarrow j}$.
Then the mean field distribution can be formulated as
\begin{align}
\label{eq:mfa}
\begin{split}
    Q(\mathbf{x}|I, B_I) & = \prod_{i = 1}^n Q(x^{cls}_i, x^{bbox}_i| h_i) Q(h_i|f^v_i)\\
    & \prod_{j \neq i} Q(x_{i \rightarrow j}| h_{i\rightarrow j}) Q(h_{i\rightarrow j}| f^e_{i\rightarrow j})
\end{split}
\end{align}
where $f^v_i$ is the visual feature of the $i$-th node, and  $f^e_{i\rightarrow j}$ is the visual feature of the edge from the $i$-th node to the $j$-th node.  
In the first iteration, the GRU units take the visual features $f^v$ and $f^e$ as input (Fig.~\ref{fig:architecture}(a)).
We use the visual feature of the proposal box as the visual feature $f^v_i$ for the $i$-th node. We use the visual feature of the union box over the proposal boxes $b_i, b_j$ as the visual feature $f^e_{i \rightarrow j}$ for edge $i \in j$. These visual features are extracted by an ROI-pooling layer~\cite{FASTRCNN} from the image. In later iterations, the inputs are the aggregated messages from other GRU units of the previous step. We talk about how the messages are aggregated and passed in the next subsection.

\subsection{Primal Dual Update and Message Pooling}
\label{sec:primal_dual}
Sec.~\ref{sec:rnn} offers a generic formulation for solving graph inference problem using RNNs. However, we observe that we can further improve the inference efficiency by leveraging the unique bipartite structure of a scene graph. In the scene graph topology, the neighbors of the edge GRUs are node GRUs, and vice versa. Passing messages along this structure forms two disjoint sub-graphs that are the dual graph to each other. Specifically, we have a node-centric primal graph, in which each node GRU gets messages from its inbound and outbound edge GRUs. In the edge-centric dual graph, each edge GRU gets messages from its subject node GRU and object node GRU (Fig.~\ref{fig:architecture}(b)). We can therefore improve inference efficiency by iteratively passing messages between these two sub-graphs instead of through a densely connected graph~(Fig.~\ref{fig:architecture}(c)).

As each GRU receives multiple incoming messages, we need an aggregation function that can fuse information from all messages into a meaningful representation.
A na\"{i}ve approach would be standard pooling methods such as average- or max-pooling. However, we found that it is more effective to learn adaptive weights that can modulate the influences of incoming messages and only keep the relevant information. We introduce a \emph{message pooling} function that computes the weight factors for each incoming message and  fuse the messages using a weighted sum. We provide an empirical analysis of different message pooling functions in Sec.~\ref{sec:experiment}. 

Formally, given the current GRU hidden states of nodes and edges $h_i$ and $h_{i \rightarrow j}$,
we denote the messages to update the $i$-th node as $m_i$, which is computed by a function of its own hidden state $h_i$, and the hidden states of its outbound edge GRUs $h_{i\rightarrow j}$ and inbound edge GRUs $h_{j \rightarrow i}$. Similarly, we denote the message to update the edge from the $i$-th node to the $j$-th node as $m_{i\rightarrow j}$, which is computed by a function of its own hidden state $h_{i\rightarrow j}$, the hidden states of its subject node GRU $h_i$ and its object node GRU $h_j$. To be more specific, $m_i$ and $m_{i \rightarrow j}$ are computed by the following two adaptively weighted message pooling functions:

\begin{align}
    &m_i   = \sum_{j: i\rightarrow j} \sigma(\mathbf{v}_1^T[h_i, h_{i\rightarrow j}]) h_{i\rightarrow j} + \sum_{j: j \rightarrow i} \sigma(\mathbf{v}_2^T[h_i, h_{j\rightarrow i}]) h_{j\rightarrow i} \label{eq:vert}\\
    &m_{i\rightarrow j}  = \sigma(\mathbf{w}_1^T[h_i, h_{i\rightarrow j}]) h_{i} + \sigma(\mathbf{w}_2^T[h_j, h_{i\rightarrow j}]) h_{j} \label{eq:edge}
\end{align}
\noindent where $[\cdot]$ denotes a concatenation of vectors, and $\sigma$ denotes a sigmoid function. $\mathbf{w}_1$, $\mathbf{w}_2$ and $\mathbf{v}_1$, $\mathbf{v}_2$ are learnable parameters. These two equations describe the primal-dual update rules, as shown in (b) of Fig.~\ref{fig:architecture}.

\subsection{Implementation Details}
\label{sec:loss}
Our final output layers follow closely with the faster R-CNN setup~\cite{renNIPS15fasterrcnn}.
We use a softmax layer to produce the final scores for the object class as well as relationship predicate. We use a fully-connected layer to regress to the bounding box offsets for each object class separately. We use the cross entropy loss for the object class and the relationship predicate. We use $\ell1$ loss for the bounding box offsets.

We use an MS COCO-pretrained VGG-16 network to extract visual features from images. 
We freeze the weights of all convolution layers, and only finetune the fully connected layers, including the GRUs. The node GRUs and the edge GRUs have both 512-dimensional input and output. During training, we first use NMS to select at most 2,000 boxes from all proposed boxes $B_I$, and then randomly select 128 boxes as the object proposals. Due to the quadratic number of edges and sparsity of the annotations, we first sample all edges that have labels. If an image has less than 128 labeled edges, we fill the rest with unlabeled edges. At test time, we use NMS to select at most 50 boxes from the object proposals with an IoU threshold of 0.3. We make predictions on all edges except the self-connections at the test time. 

\section{Experiments}
\label{sec:experiment}
We evaluate our method for generating scene graphs from images. We compare our model
against a recently proposed model on visual relationship prediction~\cite{lu2016visual}. Our goal is to analyze our model in datasets with both sparse and dense relationship annotations. We use a new scene graph dataset based on the VisualGenome dataset~\cite{VG} in our main experiment. We also evaluate our model on the support relation inference task in the NYU Depth v2 dataset. The key difference between these two datasets is that scene graph annotation is very sparse: among all possible pairing of objects, only 1.6\% of them are labeled with a relationship predicate. The NYU Depth v2 dataset, on the other hand, exhaustively annotates the support of every labeled object.
Our experiments show that our model outperforms the baseline model~\cite{lu2016visual}, and can generalize to other types of relationships, in particular support relations~\cite{nyudepth}, without any architecture change.

\vspace{5pt}
\noindent\textbf{Visual Genome} We introduce a new scene graph dataset based on the Visual Genome dataset~\cite{VG}. The original VG scene graph dataset contains 108,077 images with an average of 38 objects and 22 relationships per image. However, a substantial fraction of the object annotations have poor-quality and overlapping bounding boxes and/or ambiguous object names. We manually cleaned up per-box annotations. On average, this annotation refinement process corrected 22 bounding boxes and/or names, deleted 7.4 boxes, and merged 5.4 duplicate bounding boxes per image. The new dataset contains an average of 25 distinct objects and 22 relationships per image.
In this experiment, we use the most frequent 150 object categories and 50 predicates for evaluation. As a result, each image has a scene graph of around 11.5 objects and 6.2 relationships.
We use 70\% of the images for training and the remaining 30\% for testing.

\vspace{5pt}
\noindent\textbf{NYU Depth V2} We also evaluate our model on the support relation graphs from the NYU Depth v2 dataset~\cite{nyudepth}. The dataset contains 1,449 RGB-D images captured in 27 indoor scenes. 
Each image is annotated with instance segmentation, region class labels, and support relations between regions. We use the standard split, with 795 images used for training and 654 images for testing.  

\vspace{5pt}
\subsection{Semantic Scene Graph Generation}
\label{sec:vg_eval}
\paragraph{Setup} Given an image, the scene graph generation task is to localize
a set of objects, classify their category labels, and predict relationships between each pair
of the objects. We evaluate our model on the new scene graph dataset.
We analyze our model in three setups below.

\begin{enumerate}
\item The \textbf{predicate classification} (\textsc{Pred}\textsc{Cls}) task is to predict the predicates of all pairwise relationships of a set of localized objects.  This task examines the model's performance on predicate classification in isolation from other factors.

\item The \textbf{scene graph classification} (\textsc{SG}\textsc{Cls}) task is to predict the predicate as well as the object categories of the subject and the object in every pairwise relationship given a set of localized objects.

\item The \textbf{scene graph generation} (\textsc{SG}\textsc{Gen}) task is to simultaneously detect a set of objects and predict the predicate between each pair of the detected objects. An object is considered to be correctly detected if it has at least 0.5 IoU overlap with the ground-truth box.
\end{enumerate}

We adopted the image-wise recall evaluation metrics, R@50 and R@100, that are used in Lu \etal~\cite{lu2016visual} for all the three setups. The R@$k$ metric measures the fraction of ground-truth relationship triplets \texttt{(subject- predicate-object)} that appear among the top $k$ most confident triplet predictions in an image. The choice of this metric is, as explained in~\cite{lu2016visual}, due to the sparsity of the relationship annotations in Visual Genome --- metrics like mAP would falsely penalize positive predictions on unlabeled relationships. 
We also report per-type recall@5 of classifying individual predicate. This metric measures the fraction of the time the correct predicate is among the top 5 most confident predictions of each labeled relationship triplet. 
As shown in Table~\ref{table:pred_type}, many predicates have very similar semantic meanings, for example, \texttt{on} vs. \texttt{over} and \texttt{hanging from} vs. \texttt{attached to}. The less frequent predicates would be overshadowed by the more frequent ones during training. We use the recall metric to alleviate such an effect. 

\begin{figure}[t!]
    \begin{center}
    \includegraphics[width=1\linewidth]{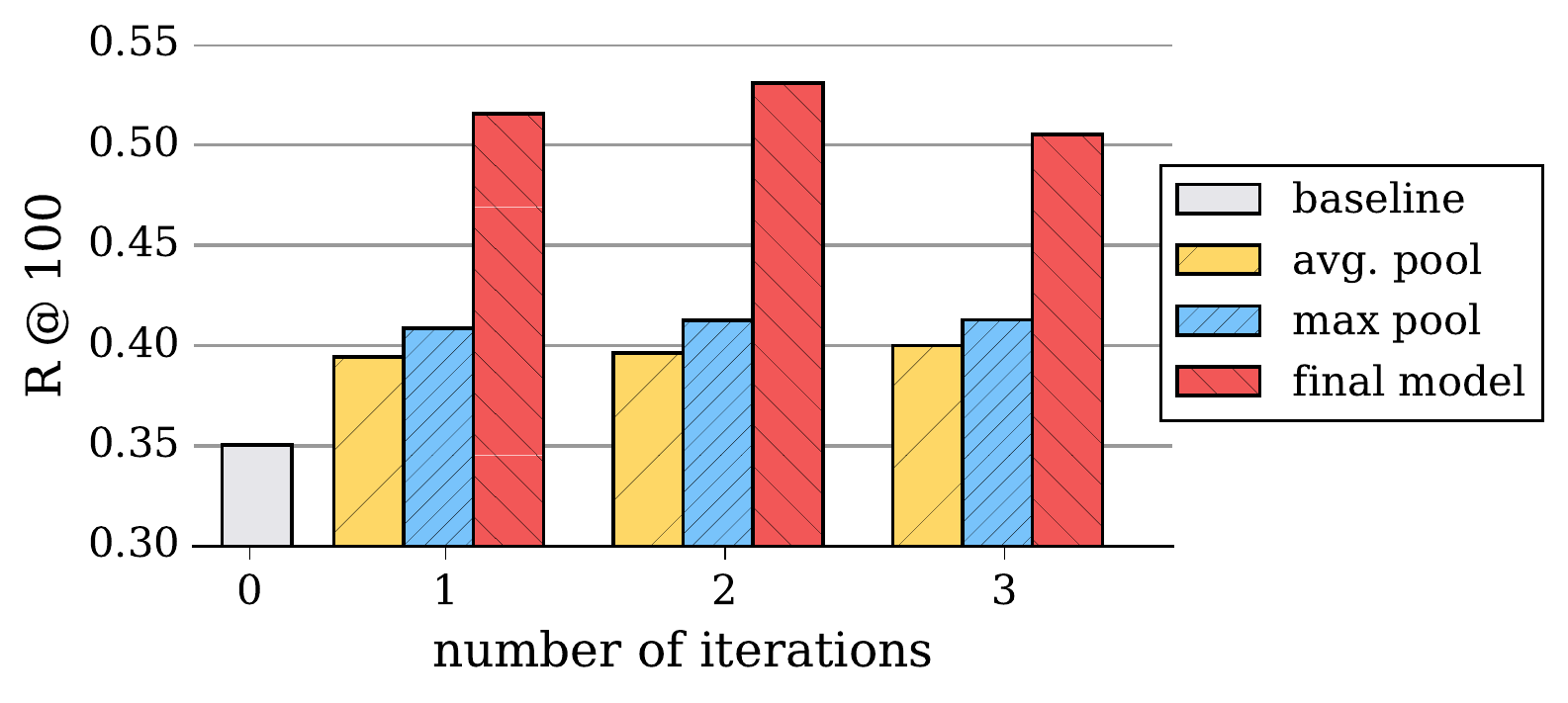}
\vspace{-20pt}
    \end{center}
\caption{Predicate classification performance (R@100) using our models with different numbers of training iterations.  Note that the baseline model is equivalent to our model with zero iteration, as it feeds the node and edge visual features directly to the classifiers.}
 \vspace{-10pt}

\label{fig:train_iter}
\end{figure}

 \vspace{-15pt}

\subsubsection{Network Models} 
\label{sec:models}
We evaluate our final model and a number of baseline models. One of the key
components in our primal-dual formulation is the message pooling functions that use learnt weighted sum to aggregate hidden states of nodes and edges into messages (see Eq.~\ref{eq:vert} and Eq.~\ref{eq:edge}). In order to demonstrate its effectiveness, we evaluate variants of our model with standard pooling methods. 
The first is to use average-pooling (\textbf{avg.~pool}) instead of the learnt weighted sum to aggregate the hidden states. The second is similar to the first one, but uses max-pooling (\textbf{max~pool}). We also evaluate our models against a relationship detection model proposed by Lu \etal~\cite{lu2016visual}. 
Their model consists of two components -- a vision module that makes predictions from images, and a language module that captures language priors. We compare with their vision module, which uses the same inputs as ours; their language module is orthogonal to our model, and can be added independently. Note that this model is equivalent to our final model without any message passing.

\begin{figure*}[t!]
\begin{center}
\includegraphics[width=1.0\linewidth]{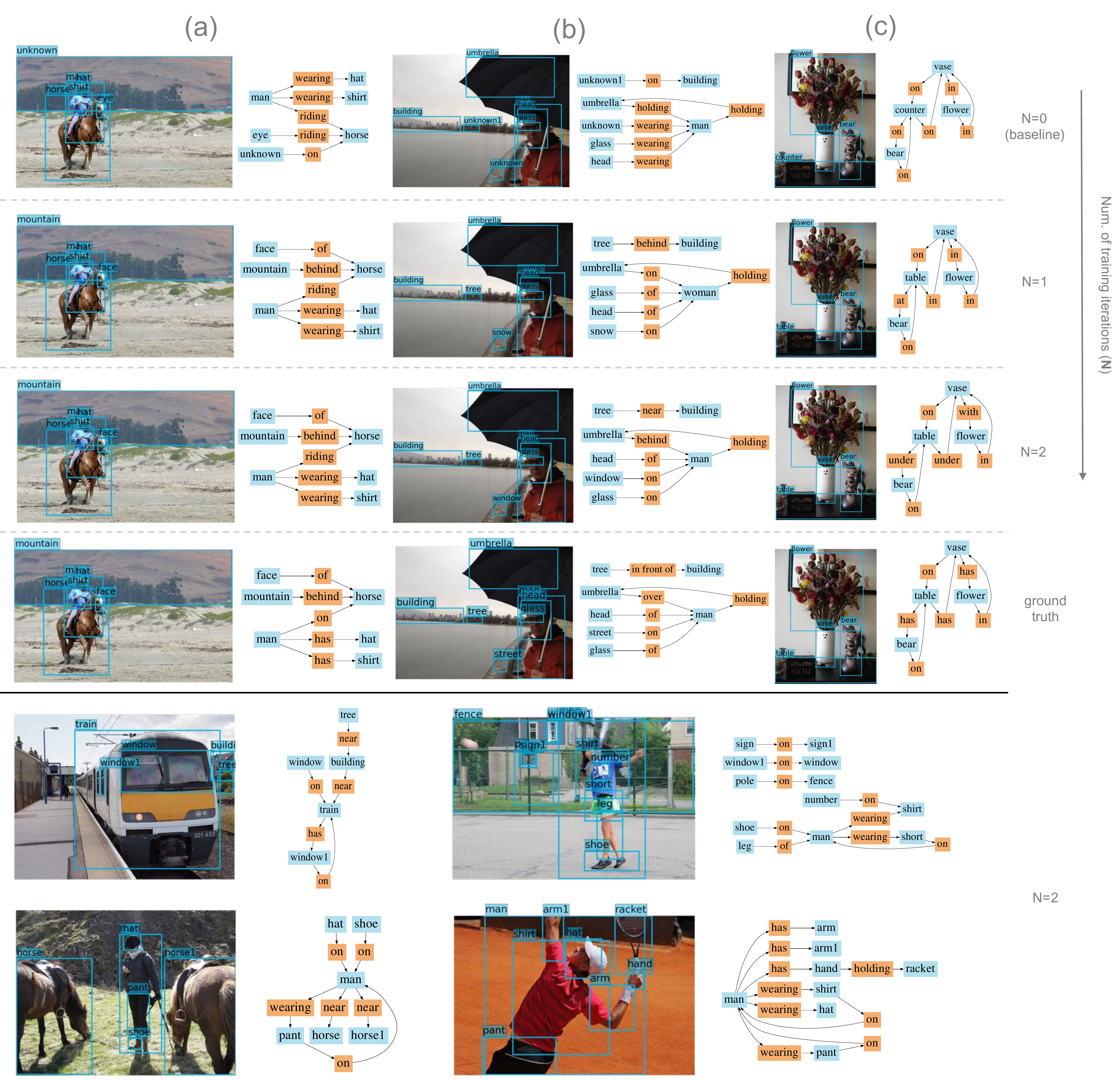}
 \vspace{-15pt}
\end{center}
\caption{Sample predictions from the baseline model and our final model trained with different numbers of message passing iterations. The models take images and object bounding boxes as input, and produce object class labels (blue boxes) and relationship predicates between each pair of objects (orange boxes). In order to keep the visualization interpretable, we only show the relationship (edge) predictions for the pairs of objects (nodes) that have ground-truth relationship annotations. }
\label{fig:qualitative}
\end{figure*}

\begin{table}[!htp]
\small
  \caption{Evaluation results of the scene graph generation task on the Visual Genome
  dataset~\cite{VG}. We compare a few variations of our model against a visual relationship detection module proposed by Lu \etal~\cite{lu2016visual} (Sec.~\ref{sec:models}).}
\vspace*{-10pt}

\begin{center}
  \begin{tabular}{l  l ||  c c c c c  }
    \multicolumn{2}{c}{}                 & ~\cite{lu2016visual} & avg. pool & max pool &final    \\ \hline\hline
    \multirow{2}{*}{\textsc{Pred}\textsc{Cls}}     &R@50  & 27.88    & 32.39    &  34.33     &\textbf{44.75} \\
                                                   &R@100 & 35.04    & 39.63    &  41.99     &\textbf{53.08} \\\hline
    \multirow{2}{*}{\textsc{SG}\textsc{Cls}}       &R@50  & 11.79    & 15.65    &  16.31     &\textbf{21.72} \\
                                                   &R@100 & 14.11    & 18.27    &  18.70     &\textbf{24.38} \\\hline
    \multirow{2}{*}{\textsc{SG}\textsc{Gen}}       &R@50  & 0.32     & 2.70     &  3.03      &\textbf{3.44}  \\
                                                   &R@100 & 0.47     & 3.42     &  3.71      &\textbf{4.24}  \\
    \hline
  \end{tabular}
 \vspace*{-10pt}

\end{center}

\label{table:vg_eval}
\end{table}

\begin{table}[!htp]
\small
  \caption{Predicate classification recall. We compare our final model (trained with two iterations) with Lu \etal~\cite{lu2016visual}. Top 20 most frequent types (sorted by frequency) are shown. The evaluation metric is recall@5.}
\begin{center}
  \begin{tabular}{c  c  c  c  c  c}
     predicate & ~\cite{lu2016visual} & ours & predicate & ~\cite{lu2016visual} & ours\\
     \hline\hline
on & \textbf{99.71} & 99.25 &under & 28.64 & \textbf{52.73} \\ 
has & \textbf{98.03} & 97.25 &sitting on & 31.74 & \textbf{50.17} \\ 
in & 80.38 & \textbf{88.30} &standing on & 44.44 & \textbf{61.90} \\ 
of & 82.47 & \textbf{96.75} &in front of & 26.09 & \textbf{59.63} \\ 
wearing & \textbf{98.47} & 98.23 &attached to & 8.45 & \textbf{29.58} \\ 
near & 85.16 & \textbf{96.81} &at & 54.08 & \textbf{70.41} \\ 
with & 31.85 & \textbf{88.10} &hanging from & 0.00 & 0.00 \\ 
above & 49.19 & \textbf{79.73} &over & \textbf{9.26} & 0.00 \\ 
holding & 61.50 & \textbf{80.67} &for & 12.20 & \textbf{31.71} \\ 
behind & 79.35 & \textbf{92.32} &riding & 72.43 & \textbf{89.72} \\ 
    \hline
  \end{tabular}
 \vspace*{-18pt}

\end{center}
\label{table:pred_type}
\end{table}

\subsubsection{Results} Table~\ref{table:vg_eval} shows the performances of our model and the baselines. The baseline model~\cite{lu2016visual} makes individual predictions on objects and relationships in isolation. The only information that
the predicate classifier takes is a bounding box covering the union of the two objects, making it likely to confuse the subject and the object. We showcase some of the errors later in a qualitative analysis.
Our final model with learnt weighted sum over the connecting hidden states greatly outperforms the baseline model ($18\%$ gain on predicate classification with R@100 metric) and the model variants. 
This shows that learning to modulate the information from other hidden states enables the network to extract more relevant information and yields superior performances.

Fig.~\ref{fig:train_iter} shows the predicate classification performances of our models trained with different numbers of iterations. The performance of our final model peaks at training with two iterations, and gradually degrades afterward. We hypothesize that this is because as the number of iterations increases, noisy messages start to permeate through the graph and hamper the final prediction. The max-pooling and average-pooling models, on the other hand, barely improve after the first iteration, showing ineffective message passing due to these na\"{i}ve aggregation methods.

Finally, Table~\ref{table:pred_type} shows results of per-type predicate recall. Both the baseline model and our final model perform well in predicting frequent predicates. However, the gap between the models expands for less frequent predicates. This is because our model uses contextual information to cope with the uneven distribution in the relationship annotations, whereas the baseline model suffers more from the skewed distribution by making predictions in isolation.

\subsubsection{Qualitative results} 
Fig.~\ref{fig:qualitative} shows qualitative results that compare our final model trained with different numbers of iterations and the baseline model. The results show that the baseline model tends to confuse about the subject and the object in a relationship. For example, it predicts \texttt{(umbrella-holding-man)} in (b) and \texttt{(counter-on-vase)} in (c). Our final model trained with one iteration is able to resolve some of the ambiguity in the object-subject direction. For example, it predicts \texttt{(umbrella-on-woman)} and \texttt{(head-of-man)} in (b), but it still predicts cyclic relationships like \texttt{(vase-in-flower-in-vase)}. Finally, the final model trained with two iterations is able to make semantically correct predictions, e.g., \texttt{(umbrella-behind-man)}, and resolves the cyclic relationships, e.g., \texttt{(vase-with-flower-in-vase)}. Our model also often predicts predicates that are semantically more accurate than the ground-truth annotations, e.g., our model predicts \texttt{(man-wearing-hat)} in (a) and \texttt{table-under-vase} in (c), whereas the ground-truth labels are \texttt{(man-has-hat)} and \texttt{(table-has-vase)}, respectively. The bottom part of Fig.~\ref{fig:qualitative} showcases more qualitative results.

\begin{table}[t!]
\small
  \caption{Evaluation results of support graph generation task. \textbf{t-ag} stands for type-agnostic and \textbf{t-aw} stands for type-aware. }
\begin{center}
  \begin{tabular}{l  c  c  c  c  }
    & \multicolumn{2}{c}{Support Accuracy} & \multicolumn{2}{c}{\textsc{Pred}\textsc{Cls}} \\ \hline
    &t-ag & t-aw & R@50 & R@100 \\ \hline\hline
    Silberman \etal~\cite{nyudepth} & 75.9& 72.6 & - & - \\
    Liao \etal~\cite{liao2016support} & 88.4 & 82.1 & - & -  \\
    Baseline~\cite{lu2016visual} & 87.7 & 85.3 & 34.1 & 50.3 \\
    Final model (ours) & \textbf{91.2} & \textbf{89.0} & \textbf{41.8} & \textbf{55.5} \\
    \hline
  \end{tabular}
\end{center}
\label{table:nyu_eval}
 \vspace*{-15pt}

\end{table}

\vspace{5pt}
\subsection{Support Relation Prediction}
We then evaluate on the NYU Depth v2 dataset~\cite{nyudepth} with densely labeled support relations. We show that our model can generalize to other types of relationships and is effective on both sparsely and densely labeled relationships. 
 \vspace*{-5pt}
\paragraph{Setup} The NYU Depth v2 dataset contains three types of support relationships: an object can be supported by an object from behind, by an object from below, or supported by a hidden object. Each object is also labeled with one of the four structure classes: \texttt{\{floor, structure, furniture, prop\}}. We define the support graph generation task as to predicting both the support relation type between objects and the structure class of each object. We take the smallest bounding box that encloses an object segmentation mask as its object region. We assume ground-truth object locations in this task.

We compare our final model with two previous models~\cite{nyudepth,liao2016support} on the support graph generation task. Following the metric used in previous work, we report two types of support relation accuracies~\cite{nyudepth}: type-aware and type-agnostic. We also report the performance with R@50 and R@100 measurements of the predicate classification task introduced in Sec.~\ref{sec:vg_eval}. Note that both \cite{nyudepth} and \cite{liao2016support} use RGB-D images, whereas our model uses only RGB images.
 \vspace*{-15pt}

\begin{figure}[t!]
\begin{center}
\includegraphics[width=1.0\linewidth]{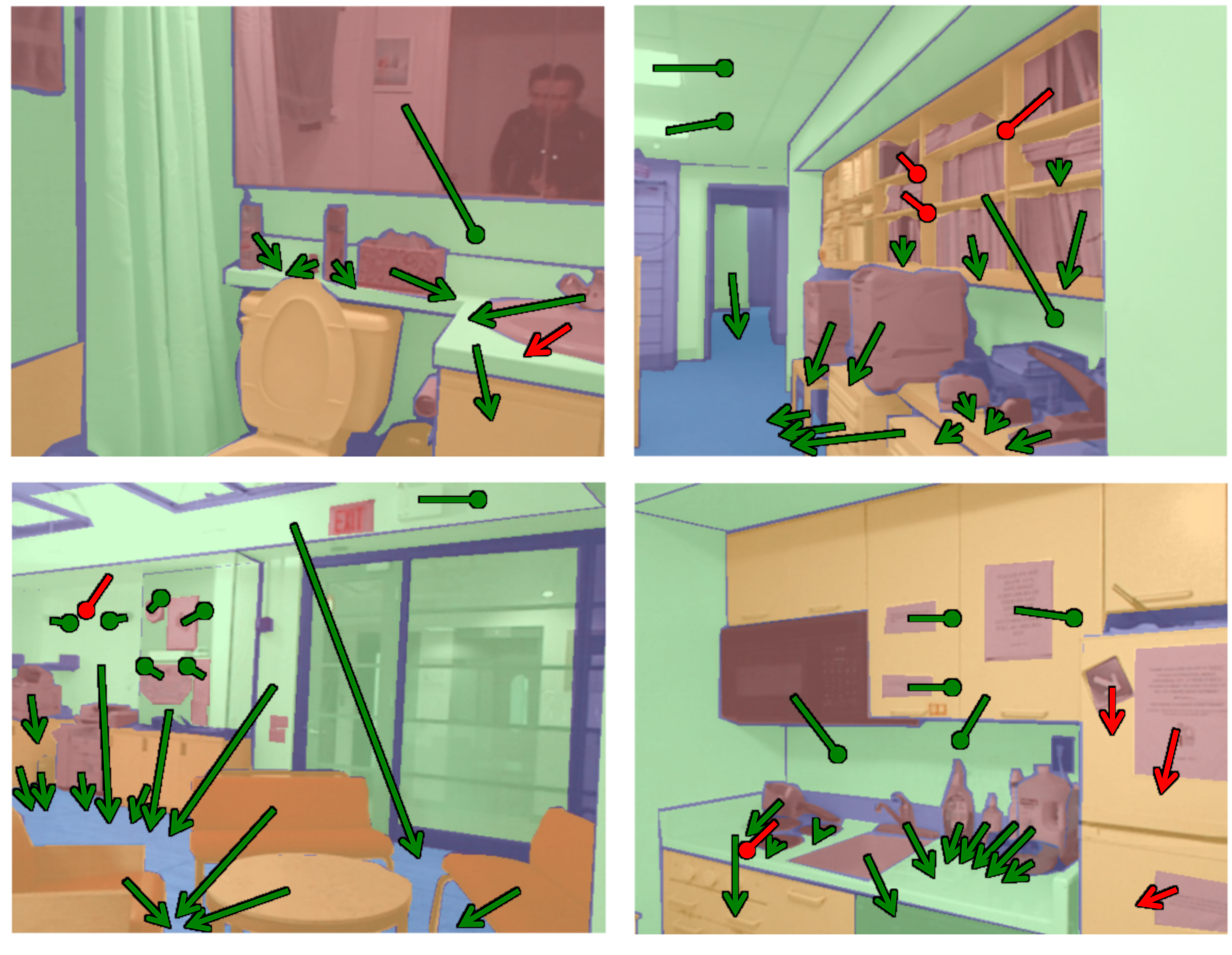}
\caption{Sample support relation predictions from our model on the NYU Depth v2 dataset~\cite{nyudepth}. $\rightarrow$: support from below, $\multimap$: support from behind. Red arrows are incorrect predictions. We also color code structure classes: ground is in blue, structure is in green, furniture is in yellow, prop is in red. Purple indicates missing structure class. Note that the segmentation masks are only shown for visualization purpose.}
\label{fig:nyu}
\end{center}
 \vspace*{-20pt}

\end{figure}

\paragraph{Results} Our model outperforms previous work, achieving new state-of-the-art performance using only RGB images. Our results show that having contextual information further improves support relation prediction, even compared to purpose-built models~\cite{liao2016support,nyudepth} that used RGB-D images. Fig.~\ref{fig:nyu} shows some sample predictions using our final model. Incorrect predictions typically occur in ambiguous supports, e.g., books in shelves can be mistaken as being supported from behind (row 1, column 2). Geometric structures that have weak visual features also cause failures. In row 2, column 1, the ceiling at the top left corner of the image is predicted as supported from behind instead of supported below by the wall, but the boundary between the ceiling and the wall is nearly invisible. Such visual uncertainty may be resolved by having additional depth information.

\section{Conclusions}

We addressed the problem of automatically generating a visually grounded scene graph from an image by a novel end-to-end model. Our model performs iterative message passing between the primal and dual sub-graph along the topological structure of a scene graph. This way, it improves the quality of node and edge predictions by incorporating informative contextual cues. Our model can be considered a more generic framework for graph generation problem. In this work, we have demonstrated its effectiveness in predicting Visual Genome scene graphs as well as support relations in indoor scenes. A possible future direction would be to explore its capability in other structured prediction problems in vision and other problem domains. 

\paragraph{Acknowledgements}  We would like to thank Ranjay Krishna, Judy Hoffman, JunYoung Gwak, and anonymous reviewers for useful comments. This research is partially supported by a Yahoo Labs Macro award, and an ONR MURI award.
{\small
\bibliographystyle{ieee}
\bibliography{main_bib}
}

\end{document}